\documentclass[letterpaper,10pt,conference]{ieeeconf}
\IEEEoverridecommandlockouts
\usepackage{amsmath,amssymb}
\usepackage{booktabs}
\usepackage{array}
\usepackage{graphicx}
\usepackage[caption=false,font=footnotesize]{subfig}
\usepackage[table]{xcolor}
\usepackage{cite}
\usepackage{url}
\usepackage[hidelinks]{hyperref}
\usepackage[final]{microtype}
\usepackage{needspace}

\newcommand{\systemname}{MonoEgo}

\title{MonoEgo: Monocular Metric Egocentric Demonstration Capture with\\
Passive Wrist Constellations and Sparse Workstation Anchors}

\author{Jie~Xu, Kangjin~Yu, Ziyi~Jin, Beichen~Wang, and Zhongpu~Xia\textsuperscript{*}\\[0.5ex]
\normalfont\normalsize Anyverse Dynamics\thanks{Jie Xu: \href{mailto:jeff_xu_0503@foxmail.com}{jeff\_xu\_0503@foxmail.com}. \textsuperscript{*}Corresponding author: Zhongpu Xia.}}

\hypersetup{pdftitle={MonoEgo: Monocular Metric Egocentric Demonstration Capture with Passive Wrist Constellations and Sparse Workstation Anchors},pdfauthor={Jie Xu, Kangjin Yu, Ziyi Jin, Beichen Wang, Zhongpu Xia}}

\begin{document}
\maketitle
\thispagestyle{plain}
\pagestyle{plain}

\begin{abstract}
Image-aligned metric demonstrations often require dedicated tracking hardware and synchronization across devices. We present \systemname{}, a capture system that replaces active wrist instrumentation with offline monocular reconstruction. One 90-FPS global-shutter camera observes calibrated passive wrist constellations, sparse workstation anchors, and the scene on a shared image clock. MonoTag SLAM combines marker corners with ORB geometry and uses visual evidence to reject ambiguous planar-marker poses. Its metric Atlas supports interval scale re-anchoring, verified map merging, and retrospective localization of earlier frames supported by the final map. Camera and wrist-constellation outputs retain validity and map provenance, and unsupported motion is left missing. Experiments show metric tracking beyond continuous anchor visibility, reconnection of supported map components, and recovery of some missing camera poses. Comparisons against a multisensor camera reference and separate stationary-constellation tests characterize trajectory agreement and precision while revealing incomplete coverage and residual geometric uncertainty. The results indicate that passive fixtures and offline reconstruction can reduce capture-side requirements. Dynamic accuracy, deployment, and downstream policy benefits require further study.
\end{abstract}

\noindent\resizebox{\columnwidth}{!}{%
\begin{tabular}{@{}l@{}}
\textbf{Code:} \url{https://github.com/jiejie567/MonoTag-SLAM}\\
\textbf{Hardware and materials:}\\
\url{https://github.com/jiejie567/MonoEgo}
\end{tabular}}

\section{Introduction}

Robot demonstrations should represent different people, homes, workspaces, and manipulation styles. Portable systems aim to extend data collection beyond robot laboratories~\cite{chi2024umi,wang2024dexcap,kareer2024egomimic}. We investigate whether one monocular camera and passive printed fixtures can recover metric camera and wrist motion with less capture infrastructure. Lower cost may broaden participation without guaranteeing dataset scale or quality.

\begin{figure}[!t]
\centering
\includegraphics[width=0.90\columnwidth]{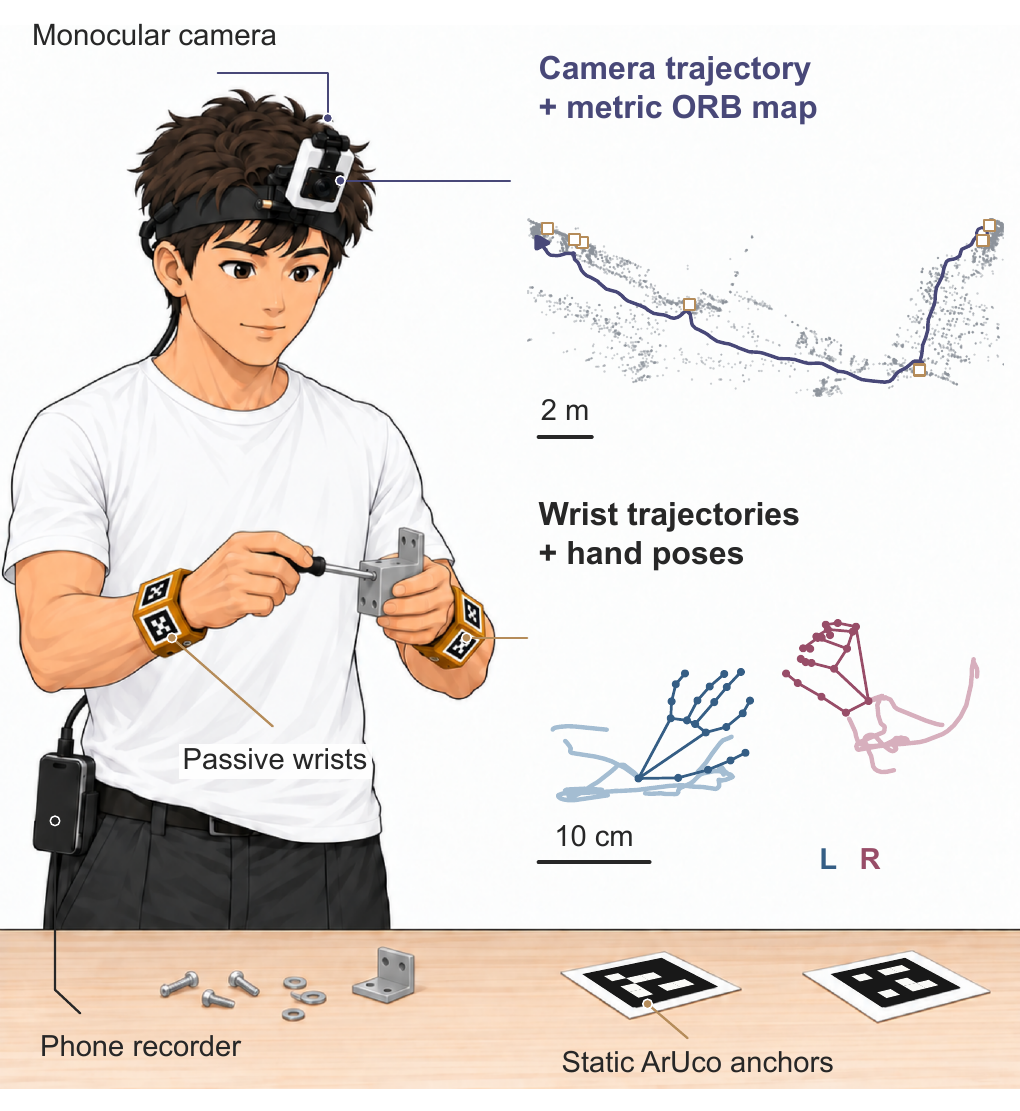}
\caption{MonoEgo capture and offline metric reconstruction. Hardware and reconstruction examples come from separate recordings.}
\label{fig:teaser}
\end{figure}

Handheld interfaces, active gloves, exoskeletons, and multisensor rigs provide complementary approaches to demonstration capture~\cite{chi2024umi,wang2024dexcap,xu2025dexumi}. Separate video and motion sensors require temporal synchronization and spatial extrinsic calibration, whether handled by the user or an integrated device. Active wearables may also add power, mass, and communication requirements. In \systemname{} (Fig.~\ref{fig:teaser}), workstation anchors, wrist markers, hands, and scene appearance share each RGB frame and camera timestamp. The wrist fixtures require no battery, inertial measurement unit (IMU), or radio. The powered camera and recorder still require camera and constellation calibration.

Simple capture hardware introduces coupled geometric uncertainties. A square marker can produce two positive-depth planar poses with similar image residuals~\cite{collins2014ippe}. Changes in the visible marker subset can also alter the preferred pose. Monocular tracking supplies multiview evidence, but does not observe absolute scale and can lose track. MonoTag SLAM treats markers and natural features as complementary constraints. It separates marker decoding from metric admission, then combines accepted corners with ORB geometry. Moving wrist markers constrain only the dynamic wrist constellations and never act as static camera anchors.

We distinguish anchored task zones from transitions between workstations. Repeated anchor observations constrain local metric geometry, so we evaluate task-zone accuracy separately from long-route continuity without claiming zero wrist drift. An unreliable transition leaves separate Atlas maps, while a new anchor can support the next task episode. Sufficient connecting evidence permits interval refinement, metric re-anchoring, or verified common-anchor merging. Marker poses remain optimizable, so anchors need not be surveyed or observed together. Task intervals retain manipulation independently of tracking success.

Offline reconstruction also separates the start of capture from the first successful localization. The final metric map can localize earlier frames and recover their wrist trajectories. \emph{Retrospective metric recovery} requires image-to-map evidence. It does not interpolate motion or rewrite the outcome of forward tracking.

This paper makes three contributions:
\begin{enumerate}
\item \textbf{Passive single-camera capture.} Calibrated printable wrist constellations and a high-rate image clock remove active wrist electronics and synchronization between image and motion streams.
\item \textbf{Ambiguity-aware marker-aided ORB localization.} Reliable corners constrain metric geometry. Visual motion and gated auxiliary marker evidence screen inconsistent planar poses and marker subsets.
\item \textbf{Metric consistency and retrospective recovery.} An unsurveyed-anchor Atlas connects supported episodes through interval refinement and verified merging, and re-localizes supported pre-initialization frames without fabricating unobserved motion.
\end{enumerate}
Validity, map revision, and immutable-RGB provenance define the output contract. Evaluation reports camera-reference error, coverage, fragmentation, retrospective recovery, and constellation precision.

\section{Related Work}

\subsection{Portable demonstration capture}
Portable capture systems trade sensing infrastructure against embodiment and deployment requirements. UMI and FastUMI use handheld grippers~\cite{chi2024umi,zhaxi2025fastumi}, while UMI-3D adds synchronized LiDAR and inertial sensing for metric 3D reconstruction~\cite{wang2026umi3d}. DexCap and DexUMI use active hand tracking or exoskeleton interfaces~\cite{wang2024dexcap,xu2025dexumi}. EgoMimic and EgoDex instead use integrated headsets for egocentric video and hand motion~\cite{kareer2024egomimic,hoque2026egodex}.

Our design requires constellation calibration and visible markers, but removes wrist electronics and cross-device synchronization. Integrated devices may already synchronize internally. No matched cost or accuracy study was performed.

\subsection{Fiducial-assisted monocular SLAM}
ORB-SLAM3 provides feature tracking, relocalization, loop closure, and a multimap Atlas~\cite{campos2021orbslam3}. LDSO adds loop closure to direct sparse odometry, while DROID-SLAM uses learned recurrent updates and dense bundle adjustment~\cite{gao2018ldso,teed2021droid}. Their monocular baselines do not receive our known-size marker constraints. UcoSLAM and sSLAM combine markers with natural or temporary keypoints~\cite{munoz2019ucoslam,romero2023sslam}; TagSLAM provides tag factors and optional external odometry~\cite{pfrommer2019tagslam}.

Infinitesimal plane-based pose estimation (IPPE) exposes the two-solution structure of planar pose estimation~\cite{collins2014ippe}. Planar-pose disambiguation and joint marker-feature optimization are not new in isolation. Our contribution combines ambiguity-aware metric admission, interval consistency, and retrospective recovery, while separating static anchors from dynamic wrist constellations and propagating validity into exported labels.

\subsection{Egocentric hand labels}
H2O, HOT3D, and EgoDex provide hand-object or manipulation data through dedicated capture pipelines~\cite{kwon2021h2o,banerjee2024hot3d,hoque2026egodex}. MediaPipe Hands and WiLoR estimate hands from images, while VibeMesh adds acoustic sensing~\cite{zhang2020mediapipehands,potamias2025wilor,mao2025vibemesh}.

HaWoR combines monocular camera and hand reconstruction with learned motion infilling~\cite{zhang2025hawor}. Our distinction is physical marker scale, calibrated rigid constellations, and geometric validity under intermittent anchors. Optional WiLoR/HaWoR components remain outside MonoTag SLAM.

\section{Accessible System and Data Workflow}

\begin{figure*}[!t]
\centering
\includegraphics[width=0.98\textwidth]{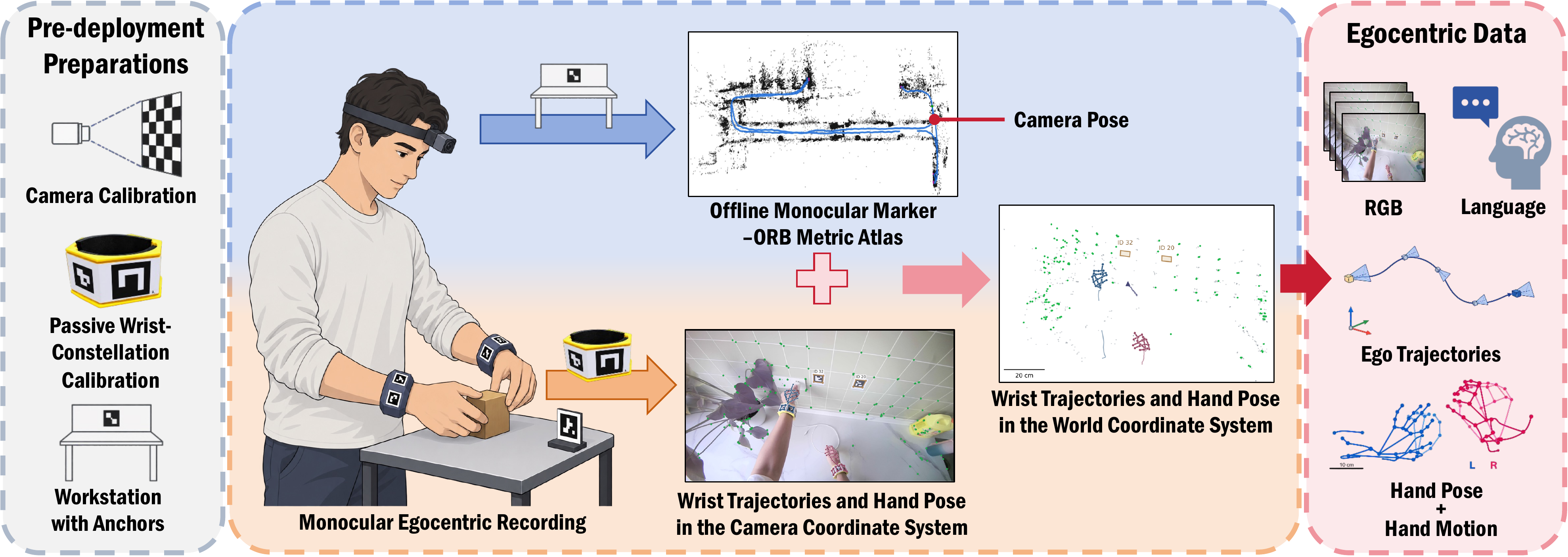}
\caption{MonoEgo workflow. Camera and wrist-constellation calibration precede monocular recording. Offline MonoTag reconstruction combines natural features, workstation anchors, and wrist observations to estimate metric camera and wrist-constellation trajectories; conditional hand estimates and frozen RGB frames form the exported egocentric data.}
\label{fig:workflow}
\end{figure*}

Fig.~\ref{fig:workflow} summarizes the pipeline from one-time calibration and monocular recording to offline metric reconstruction and exported egocentric data products.

\subsection{Capture-side design}
The kit contains one head-mounted monocular camera, two printable wrist constellations carrying ArUco markers~\cite{garrido2014aruco}, and one static anchor instance per workstation. Passive refers to the printed fixtures, not the powered camera and recorder. An anchor may contain several IDs; sparsity is counted per physical workstation instance. We record 1920$\times$1080 global-shutter RGB at 90 FPS with an 8-ms exposure, limiting motion blur and providing denser opportunities for marker detection during rapid motion. Processing remains offline; 90 FPS is the acquisition, not reconstruction, rate.

We use a WN-L2406K397L global-shutter module with a 2.3-mm $f/1.8$ M12 lens. Its nominal diagonal/horizontal/vertical fields of view are $130^\circ/117^\circ/78^\circ$; reconstruction uses calibrated intrinsics. The capture-side bill of materials is approximately US\$78--80, including the original camera module, head mount, passive fixtures, and print material. The recorder, replacement lens, fasteners, elastic bands, labour, shipping, and offline host are excluded.

\subsection{One-time constellation layout calibration}
Each assembled fixture is calibrated once because printing and fastening perturb nominal geometry. A multiview RGB sequence observes adjacent marker faces. Robust pairwise initialization rejects inconsistent planar poses; corner bundle adjustment jointly refines relative marker transforms and camera poses. A support-weighted spanning tree expresses each marker as $^{B}\mathbf{T}_{M_i}$ in a shared body frame $B$. Disconnected layouts are rejected. The reused frozen layout represents fixture geometry, not anatomy.

\subsection{Processing and output validity}
After camera and constellation calibration, each recording is stored as immutable raw RGB. Cached observations feed offline marker-aided ORB reconstruction and subsequent label export. Every output records its localization source, metric state, map identity, and revision. A world-frame constellation pose requires a valid camera pose, a metric map, and accepted fixture observations. Map corrections consistently update historical frame-to-keyframe transforms and apply each scale change once. Missing measurements are neither frozen nor interpolated, and unaligned maps are exported separately.

Conditional hand estimates and validity-aware training exports are optional, not robot commands. Marker covering or video inpainting~\cite{zhou2023propainter} acts only on derived RGB after labels are frozen; edited frames never feed localization. Appearance edits neither recover occluded content nor ensure privacy.

\section{Ambiguity-Aware Marker-Aided ORB Localization and Metric Atlas}

\subsection{Frames, variables, and gauge states}
Let $\mathbf{T}_{WC_t}\in SE(3)$ be the camera pose in Atlas world, $\mathbf{T}_{WM_j}$ a static anchor-marker pose, and $\mathbf{p}_{jk}$ marker corner $k$ in marker coordinates. For an accepted image corner $\mathbf{u}_{tjk}$, the residual is
\begin{equation}
\mathbf{r}_{tjk}=\mathbf{u}_{tjk}-\pi\!\left(\mathbf{K}\mathbf{T}_{C_tW}\mathbf{T}_{WM_j}\mathbf{p}_{jk}\right),
\label{eq:marker}
\end{equation}
with robust, quality-dependent pixel covariance. Once a map is metric, the factors in Eq.~\ref{eq:marker} enter the native pose, local bundle adjustment (BA), and global BA with ORB observations. Before metricization, metre-valued marker corners provide scale evidence but do not enter arbitrary-gauge BA. They become active projection factors only after a validated metric transform commits atomically. Atlas world and marker world then share one frame. Scale state is distinct from tracking validity: loss does not erase an existing map's scale.

For a metric optimization window, let $\mathcal{X}$ contain its free camera poses, background points $\mathbf{X}_\ell$, and eligible static-marker poses. The joint reprojection objective is
\begin{equation}
\begin{split}
\min_{\mathcal{X}}\quad &\sum_{(t,\ell)\in\mathcal{O}}
\rho_O\!\left(\|\mathbf{v}_{t\ell}-\pi(\mathbf{K}\mathbf{T}_{C_tW}\mathbf{X}_\ell)\|^2_{\Omega^O_{t\ell}}\right)\\
&+\sum_{(t,j,k)\in\mathcal{M}}
\rho_M\!\left(\|\mathbf{r}_{tjk}\|^2_{\Omega^M_{tjk}}\right).
\end{split}
\label{eq:joint-ba}
\end{equation}
\looseness=-1
Here $\mathcal{O}$ and $\mathcal{M}$ denote admitted ORB and marker observations, $\|r\|^2_\Omega=r^\top\Omega r$, and $\rho$ is the robust loss used during staged optimization. ORB information depends on pyramid scale, whereas marker information reflects observation quality. Pose-only optimization keeps landmarks fixed. Native local BA retains registered marker geometry. Staged marker-layout BA frees markers seen in at least two keyframes. Independent markers have separate rigid-pose variables and no assumed coplanarity or fixed transform between IDs. A calibrated board shares one rigid-layout transform. Physical corner geometry and marker side lengths remain fixed. Fixing the origin camera removes the rigid gauge, while admitted known-size corners constrain metric scale. Local boundary keyframes remain fixed where required. A lower residual is not sufficient for acceptance. The gauge and geometry checks described below must also pass.

For wrist-constellation body $B_t$, the calibrated layout $\{^{B}\mathbf{T}_{M_i}\}$ jointly estimates $\mathbf{T}_{C_tB_t}$ from all accepted wrist-marker corners; the reported world pose is
\begin{equation}
\mathbf{T}_{WB_t}=\mathbf{T}_{WC_t}\mathbf{T}_{C_tB_t}.
\label{eq:wrist}
\end{equation}
Wrist markers are dynamic observations and are excluded from the static map. Without a separately calibrated wearer-specific transform, $\mathbf{T}_{WB_t}$ is a wrist-constellation pose rather than an anatomical wrist pose.

\subsection{Ambiguity-aware marker admission}
Marker decoding and metric pose admission are separate decisions. Each isolated square first produces all positive-depth IPPE hypotheses. An observation remains unresolved when two candidates differ by more than $5^\circ$ or 20 mm in camera centre, while their corner root-mean-square (RMS) errors differ by less than 0.5 px. Such an observation cannot initialize a gauge independently. A recent compatible measurement may select one branch; otherwise, the frame remains uninitialized. Gated corners from a quality-rejected neighbouring marker may distinguish branches when one candidate explains them clearly. This \emph{assist-only} evidence cannot initialize the map, set scale, or enter BA at full strength.

For moving wrist constellations, the offline pass re-evaluates short-gap IPPE candidates against the committed metric camera trajectory. It uses measured constellation poses within 0.15 s before or after the candidate, but never fills a missing wrist observation.

A changing accepted marker subset can yield a low-residual pose that contradicts other visible markers. Admission therefore evaluates support across the visible set instead of trusting the selected subset alone. During a marker-assisted loop or interval proposal, an incomplete group may be excluded in at most two retries. Each retry also requires independent, complete, high-confidence views of the same keyframe and marker. Otherwise, the proposal is rolled back. This rule preserves strong marker factors, ORB loop pairs, and the gauge, while preventing one malformed partial decode from rejecting an otherwise supported update. A previous pose is never copied forward as a measurement.

Accepted corners enter staged joint BA. Full-map refinement fixes the origin keyframe while jointly refining camera poses, background points, and eligible marker poses. Reprojection, positive-depth and gauge-consistency checks precede publication of the optimized states.

\subsection{Unsurveyed-anchor initialization and metricization}
Marker admission uses border margin, projected area, incidence angle, IPPE dual-solution consistency~\cite{collins2014ippe}, and corner reprojection residual. A strong anchor observation can seed a metric camera pose while the camera is stationary. This establishes a meter-valued pose and anchor landmark, but does not invent background depth. ORB points enter only after positive-depth, parallax, and multiview triangulation checks yield at least 50 valid points. Without an anchor, ordinary monocular initialization waits for sufficient texture and baseline; it emits no pose while the head is stationary and initialization is underconstrained.

A single PnP estimate cannot metricize an existing arbitrary-scale map. A scale proposal requires at least eight reliable observations across three keyframes and approximately 4 cm of marker-estimated translation. It must also pass uncertainty and multiview-depth checks. Independently, the ORB baseline divided by median scene depth must exceed a fixed threshold. This parallax gate prevents near-zero-baseline PnP jitter from producing a large scale estimate.

Until these tests pass, a direct marker pose may be valid locally while the background map remains arbitrary-scale. If reliable anchor evidence persists for 2 s without observable visual scale, we preserve the arbitrary transition map. A separate marker-seeded map then provides local metric pose without implying that the transition map was scaled or merged.

A previously unknown static marker observed along a metric trajectory begins as a candidate world landmark. It becomes a registered anchor only after passing multikeyframe baseline, residual, positive-depth, and uncertainty checks. Two anchors need not appear in one image or have a measured relative transform. A continuous visual trajectory must provide the geometric evidence that connects them. Without such evidence, the anchors remain in separate map components.

\subsection{Retrospective metric recovery}
An initialization delay does not necessarily make all earlier observations unusable. After the final metric Atlas is committed, we distinguish two cases. Previously localized frames at arbitrary scale are resolved through corrected reference-keyframe transforms. Frames without an initial camera pose require new image-to-map localization, because a change of units alone cannot recover them.

The second pass reads the final Atlas without modifying it and matches raw-frame ORB descriptors to final landmarks. Prefix PnP recovery requires at least 30 inliers, no more than 3 px RMS, positive depth, spatial support, and connectivity to verified initialized poses. A separate short-gap pass retries later missing frames bounded by independently localized controls in the same resolved map. Every recovered pose requires image-to-map evidence. Neither pass fills long unsupported losses or connects unaligned maps.

For an accepted camera pose, Eq.~\ref{eq:wrist} maps the measured camera-relative constellation pose into the final metric frame; its already-metric translation is not rescaled. Missing wrist observations remain invalid. Each recovered label records its map revision and late availability, without rewriting forward tracking history. Showing final points from the start of a replay is distinct from this geometric recovery. We assess the added valid-label duration and recovery errors separately from ordinary BA or retrospective display.

\subsection{Interval metric consistency and re-anchoring}
Suppose anchor evidence at A establishes a reliable gauge, ORB tracking connects A to non-covisible anchor B, and multiframe observations at B close the interval. B may be newly registered or revisited after an anchor-absent transition. Closure requires eight reliable observations, two B-region keyframes, approximately 4 cm metric baseline, sufficient normalized visual parallax, and acceptable uncertainty.

Every geometrically valid closure schedules metric consistency refinement over the interval from A to B, even when the estimated scale is close to one. Continuous visibility cannot trigger repeated solutions of the same near-unit update. With A fixed, the optimizer includes the connecting keyframe path and its covisible neighbourhood. Staged BA then refines camera poses, ORB points, marker poses, and the original corner factors. The graph admits a Sim(3) correction, termed \emph{metric re-anchoring}, when the scale discrepancy exceeds 3\% and three estimated standard deviations. Otherwise, near-unit refinement may reduce residuals and register B, but does not create an artificial scale change.

More explicitly, the scale-correction stage uses world-to-camera similarities $S_t$ on the selected connected graph, initialized from the committed poses with unit scale. Its objective combines existing relative-pose edges, admitted marker projections, and B-region log-scale observations:
\begin{equation}
\begin{split}
E_S={}&\sum_{(i,j)\in\mathcal{E}}\|\log(\widehat S_{ji}^{-1}S_jS_i^{-1})\|^2_{\Omega_{ij}}\\
&+\sum_{b\in\mathcal{B}}\frac{(\log s_b+\log\widehat m)^2}{\sigma_{\log s}^{2}}
+E_{M}(S).
\end{split}
\label{eq:scale-graph}
\end{equation}
In Eq.~\ref{eq:scale-graph}, the A vertex is fixed, $\widehat S_{ji}$ retains the pre-correction graph relation, and $\widehat m$ is the validated metric-per-visual scale estimate. The inverse sign follows the world-to-camera convention, and $E_M$ is the robust marker-projection term. Existing graph edges are soft spatial constraints, not independent metric measurements. Marker geometry is fixed during this proposal stage; eligible marker poses and background points are then refined with Eq.~\ref{eq:joint-ba}. Thus, correction spans the connected interval rather than repeatedly rescaling the Atlas.

Depth ratio, positive depth, reprojection, and anchor consistency gate commits; failed proposals roll back. \emph{Marker-assisted loop closure} revisits a registered place. A marker of known size constrains scale, but cannot remove global pose drift without a known world pose or connecting trajectory.

After marker metricization, ordinary visual loop closure is constrained to $SE(3)$ throughout RANSAC, geometric refinement, covisible-window correction, and essential-graph optimization. Fixing only the final graph is insufficient because a free-scale Sim(3) seed can already distort the local window. Appearance-based loops may correct pose and close the graph, but cannot independently rescale a metric map. Scale remains free only for cross-map alignment and the separately validated metric re-anchoring operation. Before metricization, ordinary monocular loop and merge hypotheses retain their native Sim(3) treatment.

\subsection{Multimap tracking and common-anchor merge}
After tracking loss, ORB-SLAM3 first attempts motion-model tracking and relocalization. Persistent failure preserves the old map and creates a new map with its own scale state. Independently initialized Atlas maps have independent gauges. Each transform from a marker component to the world is therefore indexed by both map identity and marker component, and is never reused across unmerged maps. Re-observing the same registered component in another map generates a merge hypothesis. Two maps merge only through reliable visual overlap or repeated, verified observations of a common static anchor. A common metric anchor provides a candidate $SE(3)$ map transform, followed by joint optimization and residual checks. Different marker IDs, equal physical size, or an isolated detection are insufficient. The Atlas retains marker identity, scale constraints, map relations, and events.

\subsection{Constellation observations and foreground exclusion}
Multiple wrist-marker faces share the calibrated body frame $B$, rather than producing independent per-ID trajectories. Their differing orientations provide observation redundancy under rotation and partial occlusion. All accepted corners jointly constrain Eq.~\ref{eq:wrist}; one accepted marker can still contribute, subject to ambiguity gates. Calibration error in the layout perturbs the camera-relative constellation pose, whereas anchor error affects the upstream camera pose. The evaluation separates these sources.

Before FAST/ORB allocation, masks exclude detected hands, wrist markers, and static marker interiors from background quotas. An optional non-learning gate combines map projection, bidirectional optical flow, and multiframe motion consistency. Pending dynamic-scene ablation, it remains an engineering safeguard rather than a separate contribution.

\section{Experimental Protocol}

\subsection{Camera-reference recordings}
We evaluated three independently recorded Odin traversals with overlapping workspace routes, denoted \textbf{Odin-01}, \textbf{Odin-02}, and \textbf{Odin-03}. They contain 2798, 1694, and 1830 frames over 193.4, 117.1, and 126.5 s. The complete benchmark contains 6322 frames and 7.28 min of video, including failed intervals. Lossless, rectified 1600$\times$1296 frames use the corresponding bag intrinsics at approximately 14.46 Hz. Static ArUco IDs 20--49 have confirmed 48-mm black-square sides and unsurveyed world poses. These recordings evaluate camera localization, not hand or wrist accuracy.

Odin1 (Manifold Tech) fuses LiDAR, RGB imaging, and inertial sensing, and its SLAM mode supports loop closure~\cite{manifoldodin1,manifoldodindriver}. Direct range measurements provide metric information complementary to monocular geometry, motivating its use as a multisensor trajectory reference rather than verified ground truth. Image timestamps and a fixed camera-to-IMU extrinsic define the reference camera poses; only reference odometry is interpolated. No evaluated estimator receives odometry, IMU, stereo, or depth. No time offset, extrinsics, or marker size is fitted, and reference uncertainty is not quantified.

\subsection{Comparators and implementation}
The comparison includes marker-only reconstruction, monocular ORB-SLAM3, loose marker-to-ORB alignment, UcoSLAM, TagSLAM, LDSO, monocular DROID-SLAM, and MonoTag~\cite{campos2021orbslam3,munoz2019ucoslam,pfrommer2019tagslam,gao2018ldso,teed2021droid}. Upstream ORB-SLAM3 contains compatibility and export changes, and uses 2000 features, eight pyramid levels, and FAST thresholds of 20/7. The marker-only diagnostic estimates a co-visible layout and solves joint perspective-$n$-point (PnP) for each frame. Loose alignment robustly fits one full-sequence similarity per ORB map to this layout, without joint factors or interval correction.

Marker-only and TagSLAM share independent subpixel ArUco detections, not our map or poses. TagSLAM uses a verified corner-order adapter, default SLOW optimization and no external odometry. UcoSLAM uses its own detector with the same dictionary and dimensions; MonoTag retains its own quality pipeline. This is not an identical-detector ablation. LDSO uses uncalibrated-photometric mode because response, exposure and vignetting measurements are unavailable.

Traditional baselines run headlessly in Ubuntu 20.04 containers on an Ubuntu 24.04/Core Ultra 5 225H host; MonoTag runs natively. DROID uses one NVIDIA L20, official resizing and cropping to 392$\times$488, final optimization, and dense filling. All methods use stride one and complete their backends before scoring, subject to a one-hour timeout. Complete MonoTag wall times were 638.9/275.3/315.4~s for Odin-01--03 (3.30/2.35/2.49$\times$ video duration), including observation preparation, final optimization, labels, and replay rendering. Hardware differences preclude speed comparisons.

\subsection{Accuracy, coverage and recovery}
Each connected metric map uses one $SE(3)$ alignment. Arbitrary-scale outputs use $Sim(3)$ only in a separate trajectory-shape comparison, which does not replace our $SE(3)$ score. Position absolute trajectory error (ATE) is the three-dimensional RMS distance from the camera reference. One-second translation and rotation relative pose error (RPE) use uninterrupted dense intervals~\cite{sturm2012benchmark}. We rank maps by evaluated sample count, not error, and report availability across all maps. Independently aligned fragments do not establish complete-route consistency.

UcoSLAM/LDSO final keyframes are distinguished from dense estimates; DROID filling indicates availability, not independent per-frame tracking. Exact duplicate ORB records count once. One frozen MonoTag release is used across all recordings, and baseline outputs are unchanged. One run per method and recording yields descriptive results, not frame-level confidence intervals. Retrospective recovery excludes replay smoothing and early point-cloud display.

\subsection{Stationary-constellation pilot}
A separate pilot used four stationary-constellation recordings, SW-01--04. The recordings contain 9696 frames and 107.73 s of video from a moving 1920$\times$1080 global-shutter camera at 90 FPS with an 8-ms exposure. Both constellations remained on a desk while the camera moved. All recordings used the same Ubuntu release, existing camera intrinsics and wrist layouts, and a configured anchor size of 48 mm. We did not use a zero-motion constraint, fit calibration to individual sequences, or infer finger motion. The default quality gates retained both single-marker and multimarker wrist estimates. Scatter is the three-dimensional RMS deviation from each side's valid full-sequence mean within one metric map. Coverage uses all input frames as the denominator, and missing poses remain missing. This separate pilot evaluates conditional precision, not absolute or dynamic hand accuracy.

\section{Results}

\subsection{Camera accuracy and coverage}
Table~\ref{tab:slam} reports eight methods; five of 24 runs did not complete. Entries give position ATE and evaluated-frame coverage on the largest connected map. A dagger marks keyframe-only output, and FAIL marks no completed run. MonoTag's primary-map $SE(3)$ ATE is 0.302, 0.196, and 0.150 m, with coverage of 78.1\%, 99.5\%, and 99.2\%. Availability across all metric maps is 78.1\%, 99.5\%, and 99.3\%. Odin-02 and Odin-03 provide nearly complete metric output, whereas Odin-01 lacks complete-route tracking. One-second translation/rotation RPE is 27.5 mm/$0.191^\circ$, 20.0 mm/$0.192^\circ$, and 19.3 mm/$0.230^\circ$ on uninterrupted segments. These values measure agreement with Odin odometry, not independently verified absolute accuracy.

\begin{figure}[t]
\centering
\includegraphics[width=\columnwidth]{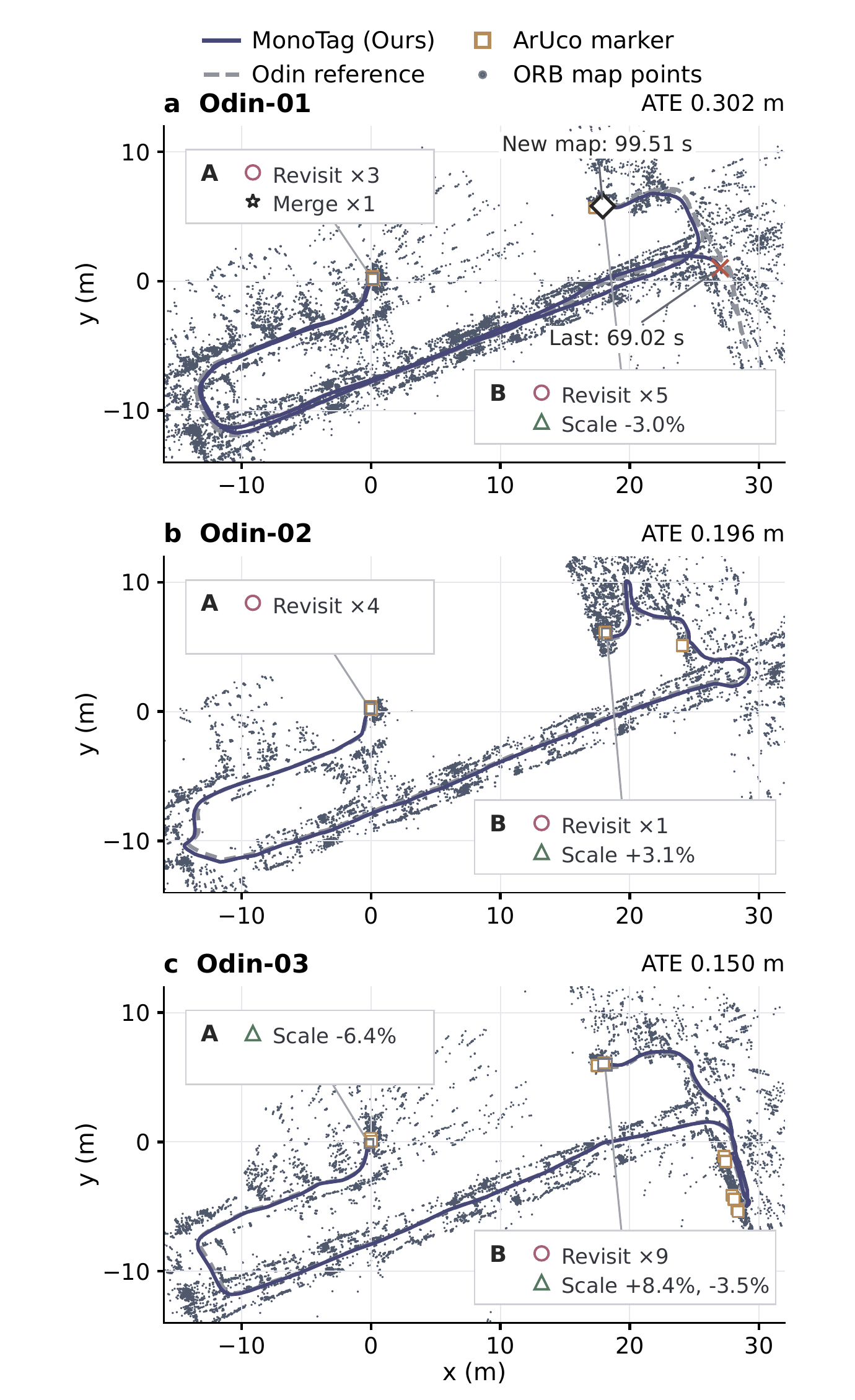}
\caption{MonoTag trajectories and sparse ORB map points after $SE(3)$ alignment to Odin, without scale fitting. Symbols mark ArUco centres, revisits, scale corrections above 3\%, map merges, and the Odin-01 gap and later submap. Only the largest connected map is shown.}
\label{fig:results}
\end{figure}

Fig.~\ref{fig:results} compares MonoTag's metric routes with Odin and locates committed marker events. MonoTag's diagnostic $Sim(3)$ ATE is 0.170, 0.106, and 0.102 m, with fitted scales of 1.019, 0.989, and 0.992. These adjustments do not replace metric scores. DROID fills every input image and obtains $Sim(3)$ ATE of 0.271, 0.316, and 1.001 m on reference-supported samples. ORB's primary-map values are 7.704, 0.113, and 0.079 m, but its output contains two, two, and four maps. The Odin-03 ORB score covers only 49.8\% of the recording.

\begin{table}[t]
\caption{Camera trajectory results on Odin recordings (ATE in m / primary-map coverage in \%).}
\label{tab:slam}
\centering\scriptsize
\setlength{\tabcolsep}{3pt}
\begin{tabular*}{\columnwidth}{@{\extracolsep{\fill}}lrrr@{}}
\toprule
Method & Odin-01 & Odin-02 & Odin-03 \\
\midrule
\rowcolor{black!6}
\multicolumn{4}{l}{\textit{Metric trajectories: $SE(3)$ alignment}} \\
Marker-only & 8.577 / 16.2 & 8.065 / 20.2 & 8.380 / 30.3 \\
ORB + loose marker & 12.271 / 86.3 & 0.800 / 89.0 & 0.250 / 49.8 \\
UcoSLAM$^\dagger$ & 0.055 / 2.8 & 0.007 / 2.2 & 0.217 / 8.2 \\
TagSLAM & FAIL & FAIL & 0.034 / 19.3 \\
MonoTag (Ours) & 0.302 / 78.1 & 0.196 / 99.5 & 0.150 / 99.2 \\
\midrule
\rowcolor{black!6}
\multicolumn{4}{l}{\textit{Arbitrary-scale trajectories: $Sim(3)$ alignment}} \\
ORB-SLAM3 & 7.704 / 86.3 & 0.113 / 89.0 & 0.079 / 49.8 \\
LDSO$^\dagger$ & FAIL & FAIL & FAIL \\
DROID-SLAM & 0.271 / 100.0 & 0.316 / 100.0 & 1.001 / 99.9 \\
MonoTag (Ours, shape) & 0.170 / 78.1 & 0.106 / 99.5 & 0.102 / 99.2 \\
\bottomrule
\end{tabular*}
\end{table}

Coverage affects marker-baseline interpretation. UcoSLAM exports 79, 37, and 150 keyframes, with forward valid-frame fractions of 15.9\%, 6.4\%, and 38.4\%. On 76, 37, and 150 shared timestamps, our $SE(3)$ ATE is 0.027, 0.008, and 0.103 m, compared with 0.055, 0.007, and 0.217 m for UcoSLAM. UcoSLAM is slightly more accurate on Odin-02's small surviving subset. On 910 shared Odin-03 timestamps, ORB's $Sim(3)$ ATE is 0.079 m versus 0.101 m for MonoTag. Greater coverage therefore need not reduce shared-support error.

TagSLAM's default SLOW optimizer raises an indeterminate-system exception on Odin-01 and Odin-02. It completes Odin-03 with an $SE(3)$ ATE of 0.034 m over 19.3\% of input frames. LDSO stops at its positive-scale check on all three recordings, so no complete trajectory is scored. These failures apply only to the reported configurations. The simple marker-only reconstruction and its loosely aligned ORB variant accumulate large layout and trajectory errors. This diagnostic does not represent every possible marker-only mapping method.

\subsection{Metric updates and retrospective recovery}
Odin-01, Odin-02, and Odin-03 contain 6, 1, and 9 metric refinements. These totals include 1, 1, and 3 scale corrections above 3\%. In the frozen benchmark, the main trajectory gap in Odin-01 spans 69.02--99.51 s. Marker 27 then seeds a new metric submap, which merges with the original map at 193.42 s. A single metric pose at 72.13 s belongs to an unmerged component. Supported components can therefore reconnect offline without recovering every intervening camera pose. Marker-revisit counts represent committed optimizations rather than independently annotated physical loop opportunities. Counts alone do not prove gains.

Final-map geometric relocalization recovers 1, 2, and 2 short-gap frames in Odin-01, Odin-02, and Odin-03 without interpolation. Controlled trials that delayed marker availability in software added 18 and 44 pre-initialization poses on Odin-02 and Odin-03. On the current Ubuntu Odin-03 pair, valid output increased from 554 to 598 of 600 frames, while all 554 shared poses remained unchanged. The 44 recovered poses have 8.78-mm RMS error against Odin after $SE(3)$ alignment fitted only to forward-valid frames. These trials establish earlier pose availability under software-withheld observations, not gains under natural occlusion or over complete routes.

\begin{figure}[t]
\centering
\includegraphics[width=0.72\columnwidth]{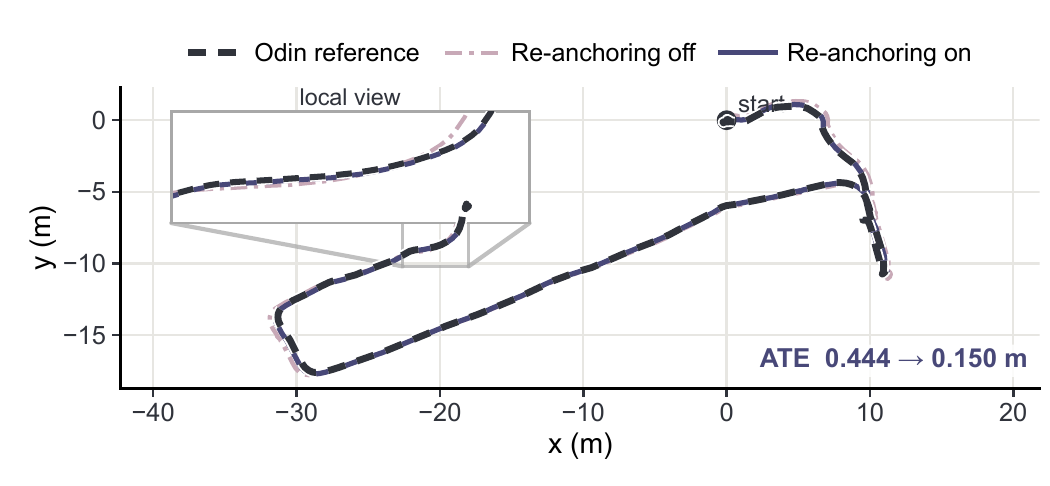}
\caption{Odin-03 trajectories with and without interval re-anchoring after $SE(3)$ alignment to the Odin reference.}
\label{fig:reanchor_ablation}
\vspace{4pt}
\end{figure}

The interval re-anchoring ablation fixed observations, calibration, and evaluation. On Odin-03, enabling versus disabling updates gave $SE(3)$ ATE/coverage of 0.150 m/99.18\% and 0.444 m/98.96\% (Fig.~\ref{fig:reanchor_ablation}); diagnostic $Sim(3)$ ATE was 0.102 versus 0.185 m. The Odin-02 effect was negligible: 0.1962 versus 0.1966 m ATE and 99.47\% versus 100.00\% coverage. The benefit is therefore data dependent, with one recording per condition.

Targeted ambiguity tests isolate two safeguards. On a controlled fixed-marker subset switch, the consistency veto reduced a reproduced camera-centre step from 83.3 to 1.63 mm; on a stationary clip it rejected none of 983 candidates. Removing auxiliary candidate hints on Odin-02 left ATE/support unchanged (0.1962 m; 1685/1694), showing that they are not universally active. These tests do not establish general marker--ORB interaction benefits.

\subsection{Constellation stability and scope}
\begin{figure}[t]
\centering
\includegraphics[width=\columnwidth]{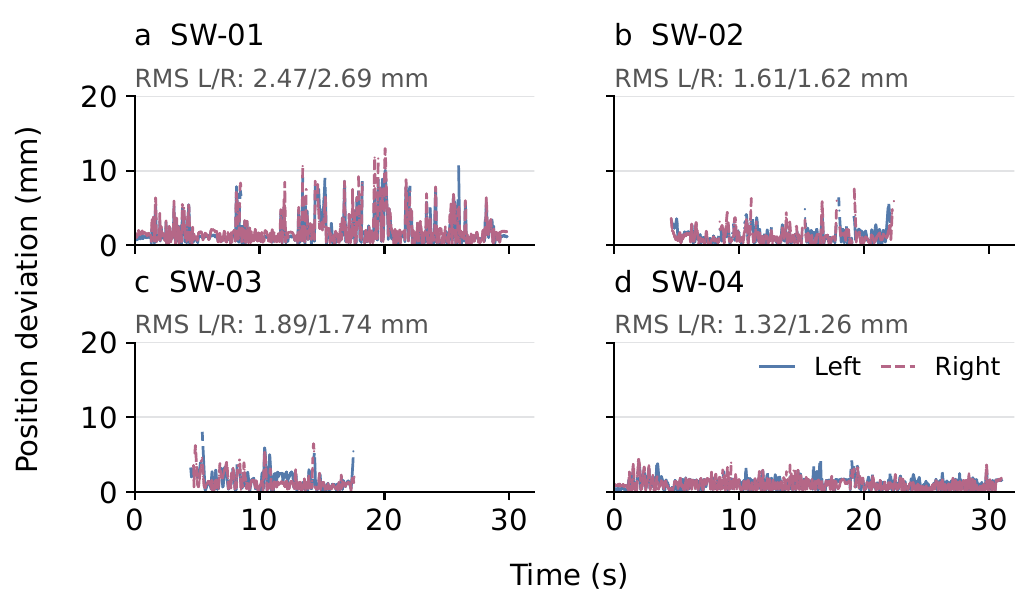}
\caption{Stationary wrist-constellation precision with a moving camera. Curves show three-dimensional displacement from each side's valid sequence mean; missing poses remain gaps.}
\label{fig:static_pilot}
\end{figure}

Fig.~\ref{fig:static_pilot} shows conditional RMS position scatter from 1.26 to 2.69 mm across eight sequence-side pairs. The largest displacement between consecutive valid input frames is 5.13 mm. Including endpoints separated by missing frames increases the maximum to 7.89 mm. The reported scatter therefore characterizes precision on accepted observations rather than uninterrupted output. Neither zero-motion constraints nor interpolated replay trajectories enter the score.

SW-01 remained in marker-seeded metric localization without a background point cloud, so it does not test ORB tracking through anchor occlusion. These four recordings do not isolate camera-speed or visibility effects, and they establish neither anatomical wrist accuracy nor dynamic joint accuracy. We did not evaluate non-expert usability or downstream learning.

\subsection{Egocentric demonstration capture}
\begin{figure}[t]
\centering
\subfloat[EGO-01]{%
  \includegraphics[width=0.485\columnwidth]{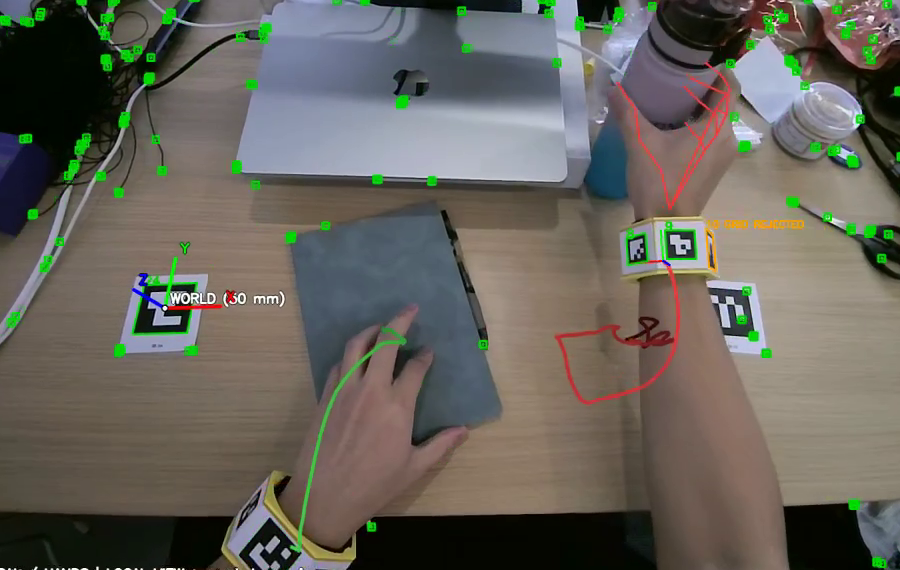}%
  \label{fig:ego01}}
\hfill
\subfloat[EGO-05]{%
  \includegraphics[width=0.485\columnwidth]{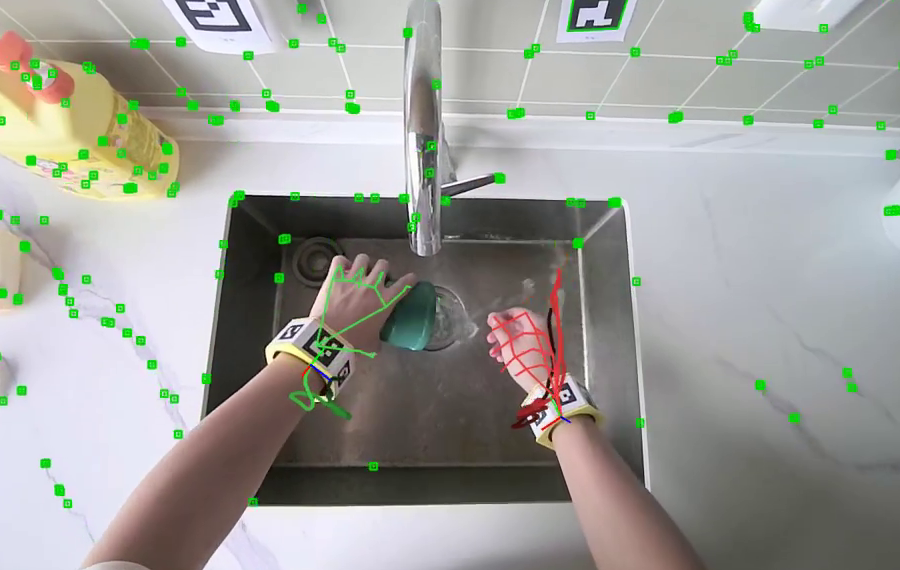}%
  \label{fig:ego05}}
\vspace{2pt}

\subfloat[Retrospective metric recovery]{%
  \includegraphics[width=\columnwidth]{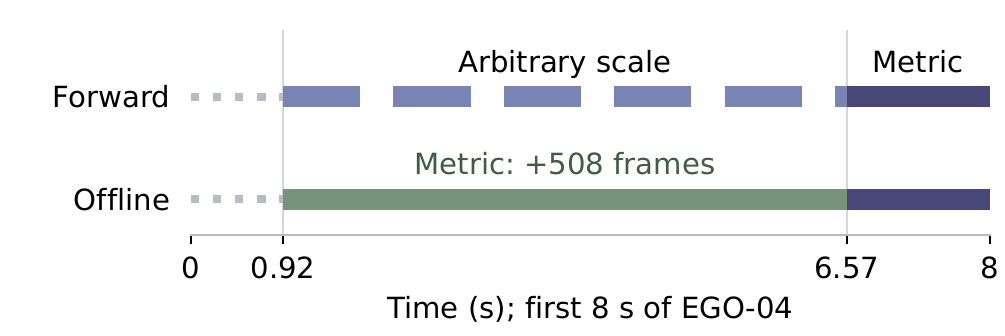}%
  \label{fig:ego_recovery}}
\caption{Egocentric outputs and offline recovery. (a, b) Annotated RGB with ORB features, marker observations, and hand overlays when available. (c) EGO-04 poses recovered retrospectively in metric units (green); dotted intervals remain unlocalized.}
\label{fig:ego_evidence}
\end{figure}

Five EGO sequences illustrate reconstruction of single-camera interactions into image-aligned camera and wrist-constellation trajectories, with hand pose when available. Figs.~\ref{fig:ego_evidence}(a) and~\ref{fig:ego_evidence}(b) show representative first-person outputs with the ORB and marker observations used by the reconstruction. Interactive replays jointly show annotated RGB and Atlas views.

\textit{Retrospective metric recovery.} EGO-04 first established metric localization at 6.57 s. Offline reconstruction recovered 508 earlier tracked frames (5.64 s), extending metric output to 0.92 s (Fig.~\ref{fig:ego_evidence}(c)); it did not invent unlocalized motion, so 0--0.92 s remained invalid.

\section{Discussion and Broader Impact}
The traversals produce metric trajectories beyond continuously visible anchors, but Odin-01 remains incomplete and some shared-support comparisons favour a baseline; replay smoothness and event counts do not establish accuracy. Printable fixtures and released tools may reduce capture barriers, but adoption remains untested and offline computation, quality control, and setup remain costs. A new anchor can support a metric task episode without recovering its route; unmerged maps cannot support globally continuous tasks. Limitations include calibration error, anchor rigidity, low texture, and incomplete hand observations. Anatomical and dynamic hand accuracy need further validation. Household RGB requires consent and governance; marker suppression does not anonymize it.

\section{Conclusion}
\systemname{} combines passive wrist constellations and one image clock with ambiguity-aware MonoTag SLAM. Across three traversals totalling 7.28 min, $SE(3)$-aligned camera position ATE against Odin odometry is 0.150--0.302 m, with 78.1--99.5\% primary-map coverage. Four stationary-constellation recordings show 1.26--2.69 mm conditional position scatter, measuring precision rather than dynamic accuracy. Potential capture-barrier reductions do not establish dynamic accuracy, usability, or policy benefit.

\section*{Acknowledgment}
OpenAI ChatGPT/Codex assisted with language editing, literature organization, and the schematic in Fig.~\ref{fig:teaser}. The authors verified all retained content.

\bibliographystyle{IEEEtran}
\bibliography{references}

\end{document}